\documentclass[journal]{IEEEtran}
\usepackage[]{graphicx}
\usepackage[abs]{overpic}
\usepackage{amsmath}
\usepackage{amsfonts}
\usepackage{amssymb}
\usepackage{multirow}
\usepackage{multirow}
\usepackage{color}
\ifCLASSINFOpdf
\else
\fi
\begin{document}
%
\title{Image-Goal Navigation in Complex Environments via Modular Learning}
%
%
%

\author{Qiaoyun~Wu$^{1}$, Jun~Wang$^{2}$,  Jing~Liang$^{3}$, Xiaoxi~Gong$^{2}$, and Dinesh~Manocha$^{3}$


\thanks{$^{1}$Q.~Wu is with the School of Artificial Intelligence, Anhui University, 230601 China.}
\thanks{$^{2}$J.~Wang, and X.~Gong are with the College of Mechanical and Electrical Engineering, Nanjing University of Aeronautics and Astronautics, 210016 China.}
\thanks{$^{3}$D. Manocha and J.~Liang are with the Department of Computer Science, the University of Maryland, College Park,  MD 20742 USA.}

}

%
%

\markboth{IEEE ROBOTICS AND AUTOMATION LETTERS, SUBMISSION}%
{Wu \MakeLowercase{\textit{et al.}}: Image-Goal Navigation in Complex Environments via Modular Learning}
%



\maketitle


\begin{abstract}
We present a novel approach for image-goal navigation, where an agent navigates with a goal image rather than accurate target information, which is more challenging. Our goal is to decouple the learning of navigation goal planning, collision avoidance, and navigation ending prediction, which enables more concentrated learning of each part.
This is realized by four different modules.
The first module maintains an obstacle map during robot navigation.
The second predicts a long-term goal on the real-time map periodically, which can thus convert an image-goal navigation task to several point-goal navigation tasks.
To achieve these point-goal navigation tasks, the third module plans collision-free command sets for navigating to these long-term goals.
The final module stops the robot properly near the goal image.
The four modules are designed or maintained separately, which helps cut down the search time during navigation and improve the generalization to previously unseen real scenes. We evaluate the method in both a simulator and in the real world with a mobile robot. The results in real complex environments show that our method attains at least a $17\%$ increase in navigation success rate and a $23\%$ decrease in navigation collision rate over some state-of-the-art models.
\end{abstract}


%
\IEEEpeerreviewmaketitle

\section{Introduction}
\begin{figure}[thpb]
\begin{center}
\includegraphics[width=\linewidth]{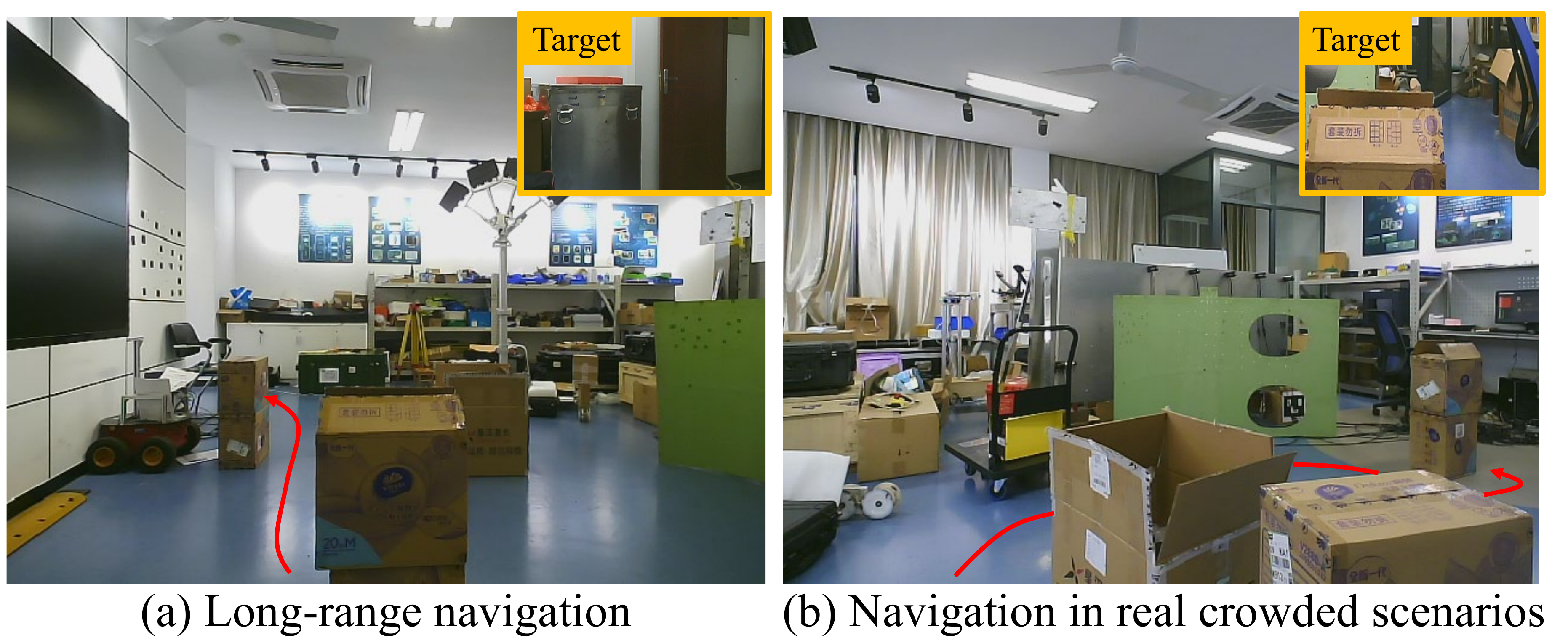}\vspace{-12pt}
\end{center}
\caption{The challenging image-goal navigation.}
\label{fig:cha_nav}\vspace{-10pt}
\end{figure}

\vspace{-4pt}
Target-driven navigation in unstructured environments remains an open problem in the robotics community. This problem is challenging, especially in settings where it is necessary to proceed without accurate target position information and with only a goal image. Furthermore, if the image-goal navigation is long-range and occurring in crowded scenarios (see Figure~\ref{fig:cha_nav}), the agent needs to learn an effective exploration strategy and a robust collision avoidance module in addition to the navigation policy. The ability of a robot to bypass obstacles safely and navigate to specified goals efficiently without a preset map, would have a great impact on robotic applications, including surveillance, inspection, delivery, and cleaning. However, although the navigation problem has been well studied in robotics and related areas for several decades~\cite{cadena2016past,zeng2020survey,WahidSCIT20}, the mobility of robots is still limited.

Most existing image-goal navigation approaches use end-to-end learning to tackle this problem ~\cite{ye2018active,pathak2018zero}. These methods combine deep convolutional neural networks (CNNs) with reinforcement learning (RL) to manage the relationship between vision and motion in a natural way. These map-less methods have presented encouraging results for image-goal navigation, and in the meantime, have shown a great tendency to overfit in the domain in which they are trained. Therefore, it is always necessary to train, or at least fine-tune, these methods for new navigation targets or environments. In addition, several recent works learn navigation policies in maze-like environments~\cite{jaderberg2016} or synthetic indoor scenes~\cite{zhu2017}, which are both much smaller and less complex than real indoor environments. Directly transferring these trained policies to real environments can be extremely challenging and impractical.

This difficulty has motivated one important set of works~\cite{chaplot2020learning,chaplot2020object,beeching2020learning}, which feature hierarchical planners combined with RL. These works insist on building spatial maps~\cite{chaplot2020learning} or topological representations~\cite{chaplot2020neural} of free space combined with a local RL-based policy executing navigation at run-time. The hierarchical and modular fashion leverages the regularities of real-world layouts, resulting in competitive performance over both geometry-based methods and recent learning-based methods. However, as we increase the complexity of the problems by requiring navigation in real scenarios with many obstacles, these RL-based methods become harder to train and do not consistently generalize well to new environments.

In this paper, we investigate alternative formulations of employing learning for image-goal navigation that does not suffer from the drawbacks of end-to-end learning while transferring well to real crowded scenarios. Our key conceptual insights lie in leveraging structural regularities of indoor environments for long-range navigation and learning a local sensor-to-controls planner for reactive obstacle avoidance. This motivates the use of learning in a modular and hierarchical fashion.
More specifically, our proposed image-goal navigation architecture comprises of an automatic mapping module, a long-term goal prediction module, a reactive motion planning module, and a robust navigation ending prediction module.
The mapping module builds explicit obstacle maps to maintain episodic memory via a learning-based system (e.g., Active Neural SLAM~\cite{chaplot2020learning}) or a geometric-based method (e.g., Gmapping~\cite{grisetti2007improved}).
The long-term goal prediction module consumes the maps with agent navigation tasks and employs learning to exploit structural regularities in layouts of real-world environments to produce distant goals for navigation.
These distant goals are used to explore the environment efficiently and finally approach the positions of navigation targets over the maps.
The motion planning module uses learning to transfer raw sensor inputs to a collision-free steering command for navigating to a long-term goal~\cite{fan2018crowdmove}.
The navigation ending prediction module learns to distinguish between the current observation and the navigation target observation, which are temporally close to or distant from one another. When the two observations are close, the navigation will be ended.
The use of mapping during image-goal navigation provides a feasible way to exploit regularities in layouts of real-world environments.
Learned long-term goal prediction can support long-range indoor navigation, while learned motion planning policies can use sensor feedbacks to exhibit effective and safe navigation behaviors.
The navigation ending judgement stops an agent properly near the image goal.

In summary, our contributions are as follows: (1) We present a hierarchical framework for image-goal navigation in real crowded scenarios. Hierarchical decomposition decouples the learning of navigation planning, collision avoidance, and navigation ending prediction, so that each part can demonstrate more concentrated learning capabilities.
(2) We propose combining long-range planning with local motion control for image-goal navigation.
This drives the robot towards the goal image, while avoiding some static or dynamic obstacles in the scenes.
(3) We demonstrate that, in simulation, our approach can significantly cut down the search time during navigation, leading to state-of-the-art navigation performances. It can also more easily transfer from simulation to the real world, including a $17\%$ increase in navigation success and a $23\%$ decrease in navigation collision overall compared with~\cite{wu2021reinforcement}, while maintaining good performances despite increasing obstacles.


The remainder of this paper is organized as follows. We review the relevant background literature in Section~\ref{sec:related}. Section~\ref{sec:Hierarchical} describes the image-goal navigation problem and presents the proposed hierarchical framework for solving the problem.
Section~\ref{sec:Experiments_and_Results} provides an exhaustive experimental validation of our approach. We conclude in Section~\ref{sec:conclu} with a summary and a discussion of future work.


\section{Related Works}
\label{sec:related}

\subsection{End-to-End Navigation System}

We focus on image-goal navigation in novel indoor environments, where no target position is available, except for the target image. Learning-based approaches commonly use end-to-end reinforcement or imitation learning for training image-goal navigation policies, which do not build a geometric map of the area. Instead, they learn the direct mapping from visual inputs to motion.
The work of~\cite{zhu2017} is the first to address the image-goal navigation problem, which designs scene-specific layers to capture the layout characteristics of a scene.
Yang et al.~\cite{yang2018visual} extend this work by incorporating a graph convolutional network into a deep reinforcement learning framework for encoding semantic priors, which shows great improvements in generalization to unseen synthetic scenes.
Wortsman et al.~\cite{wortsman2019} propose a meta-reinforcement learning navigation approach, which allows an agent to automatically learn in testing environments while completing navigation tasks effectively.
Mousavian et al.~\cite{mousavian2019visual} use a deep network to learn the navigation policy based on semantic segmentation and detection masks of visual observations.
Wu et al.~\cite{wu2021reinforcement} enhance the cross-target and cross-scene generalization of target-driven visual navigation by introducing an information-theoretic regularization term into an RL objective.
In the above methods, the navigation models need to learn navigation planning, obstacle avoidance, and scene layouts implicitly and simultaneously, which is extremely challenging and can easily result in memorizing object locations and appearance in training environments. Consequently, they always suffer from poor generalization in the real world.

\begin{figure*}[thpb]
\begin{center}
\includegraphics[width=.9\linewidth]{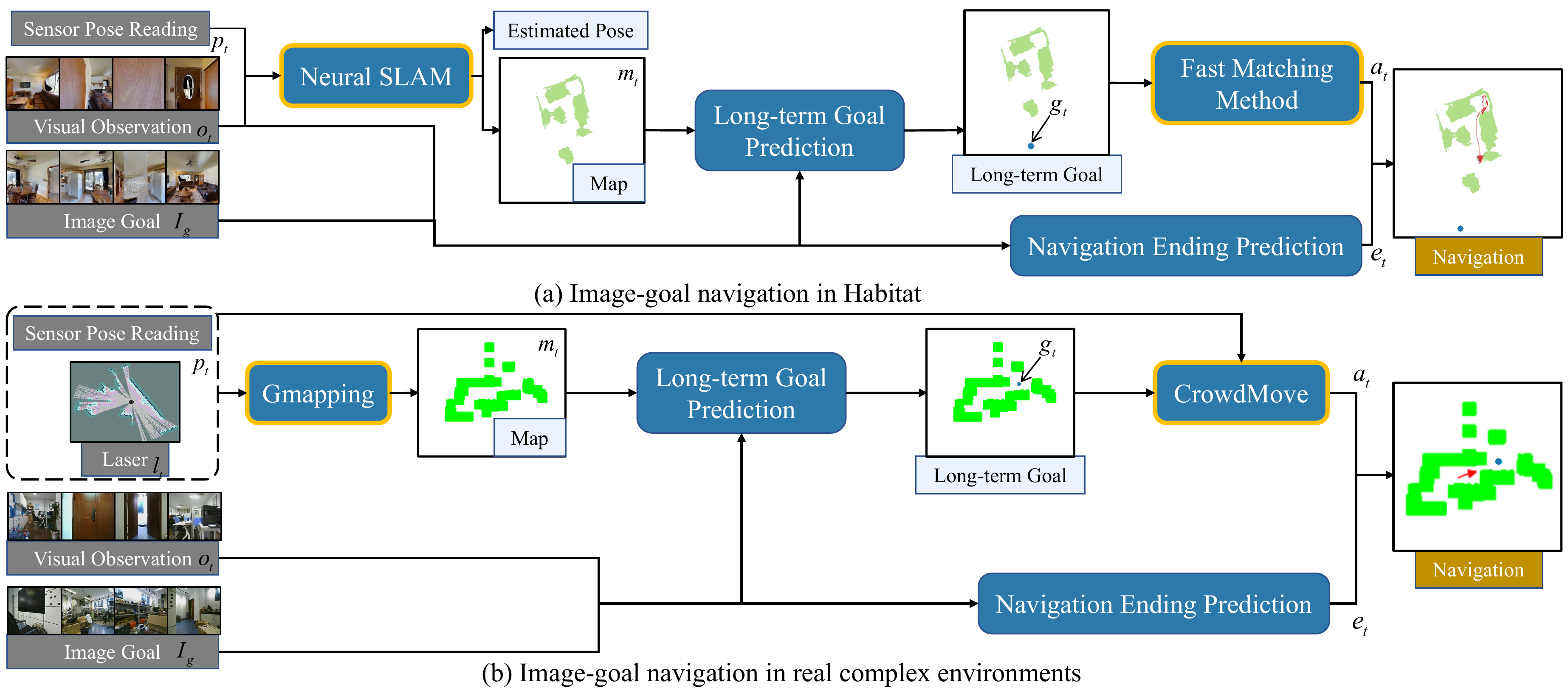}\vspace{-12pt}
\end{center}
\caption{Image-goal navigation method overview. (a) In Habitat, at each time step $t$, the Neural SLAM module~\cite{chaplot2020learning} predicts a map $m_t$ and an agent pose based on the visual observation $o_t$ and the sensor reading $p_t$. When $t$ reaches a time scale, the map $m_t$, the observation $o_t$ and the goal image $I_g$ are used by a long-term goal prediction module to output a distant goal, which is then converted to a set of navigation actions by using the Fast Marching Method~\cite{sethian1996fast}. A navigation ending prediction module is trained to stop the agent properly. (b) In real scenarios, the Gmapping~\cite{grisetti2007improved} is used to provide the online map $m_t$ and the CrowdMove~\cite{fan2018crowdmove} is used for the motion planning to long-term goals. The main differences are highlighted by orange and
the color difference of the map is to distinguish between the simulation and the real world.}
\label{fig:overview}\vspace{-10pt}
\end{figure*}

\subsection{Hierarchical and Modular Navigation System}
Hierarchical and modular navigation is an active area of research, aimed at exploiting hierarchies to navigate an agent in a scene.
There has been a recent interest in using techniques from deep learning in a modular and hierarchical fashion inside navigation systems.
For example, Chaplot et al.~\cite{mezghani2020learning} propose a fully-differentiable model for active global localization.
Devo et al.~\cite{devo2020towards} design a novel architecture composed of two main deep neural networks. The first one explores the environment and approaches the target, while the second one aims at recognizing the specified target in the view of an agent.
The work is exclusively trained in maze simulation, which is considerably simpler than real scenes.
Other works design deep models to reason spatial representations, semantic representations or topological representations, based on which navigation tasks can be executed.
The work of~\cite{chaplot2020learning} presents a Neural SLAM module for free spatial space mapping and shows strong generalization while achieving point-goal navigation tasks.
~\cite{chaplot2020object} extends the proposed model to complex semantic tasks, such as object-goal navigation, by designing a semantic Neural SLAM module that captures semantic properties of the objects in the environment.
While our work builds on the existing literature~\cite{chaplot2020learning} on spatial metric maps, the resemblance is only at the map structures. Our work focuses on balancing between scene exploration and image-goal navigation.
Chaplot et al.~\cite{chaplot2020neural} study the problem of image-goal navigation by building topological representations of space that leverage semantics and afford approximate geometric reasoning.
Beeching et al.~\cite{beeching2020learning} design a hierarchical model for long-term image-goal navigation under uncertainty in novel environments, which combines high-level graph-based planning with a local point goal policy.
Mezghani et al.~\cite{mezghani2020learning} propose a novel three-stage algorithm for learning to navigate to image goals using only RGB vision. The first stage focuses on learning visual representations and the second one explores the environment to maintain a scene memory buffer module, based on which the third stage guides an agent along a shortest path by which the agent is likely to succeed.

However, the above methods learn about navigation planning and collision avoidance simultaneously, which can be prohibitively expensive. Additionally, these methods do not consider navigation in real-world scenarios, which are characterized by heavy clutter and noise.
In contrast, we consider navigation and collision avoidance separately in the real world, and our design is inspired by the work of~\cite{francis2020long}, which proposes a hierarchical planning method for long-range point-goal navigation that combines sampling-based path planning with an RL agent as a local controller for solving obstacle avoidance.
In contrast, we achieve this by designing a global planner for long-term goal prediction followed by a local controller for reactive obstacle avoidance. This strategy has never before been applied to the image-goal navigation.

\section{Hierarchical Image-goal Navigation}
\label{sec:Hierarchical}
In the image-goal navigation, the objective is to learn a navigation controller, which can produce the shortest sequences of actions for an agent to navigate to specified goal images in a fixed time budget. The agent is initialized at a random location in the environment and receives the goal image $I_g$ as input. At each time step $t$, the agent receives visual observations $o_t$ and sensor pose readings $p_t$ and takes navigational actions $a_t$ from the controller.
In our experimental setup, all images are panoramas, including agent observations and goal images.

To tackle the image-goal navigation problem, we propose a hierarchical framework consisting of four components including an automatic mapping module, a long-term goal prediction module, a reactive motion planning module, and a robust navigation ending prediction module.
The mapping module builds a geometric map over time,
and the goal prediction module produces a long-term goal based on the map to approach the given goal image efficiently.
The motion planning module navigates an agent to the long-term goal while avoiding environmental obstacles. The navigation ending module stops the agent properly near the goal image.
Figure~\ref{fig:overview} provides an overview of the proposed framework.

\subsection{The Automatic Mapping Module}
Our mapping module is strongly decoupled from the construction and training of the rest modules.
To enable an agent to do goal planning based on an online map, we propose using some state-of-the-art SLAM methods (e.g., ORB-SLAM2~\cite{mur2017orb}, Neural-SLAM~\cite{chaplot2020learning}, or the laser-based SLAM~\cite{grisetti2007improved}) to produce the map. In the experiments, ORB-SLAM2 does not detect enough reliable keypoints in low-texture scenes, prompting a greater likelihood of lost tracking.
Considering that we must use simulators for the long-term goal prediction learning, we use the pre-trained Neural-SLAM to produce an online map $m_t$ at each time step $t$ during navigation in simulation.
However, Neural-SLAM can not bridge the simulation to reality gap, especially when the real robot setting is as simple as ours.
Hence, in real crowded scenarios, we use Gmapping~\cite{grisetti2007improved} to provide the online map and we show that the map built from the laser-based SLAM helps the agent perform as well as it does in simulation.

\subsection{The Long-term Goal Prediction Module}
\label{sec:Long-term}
In this section, we focus on learning an RL-based policy for predicting a long-term goal over an online maintained map to seek a balance between exploring an environment and approaching a navigation goal.
Our problem is, given a goal image $I_g$, at each time scale $T$, the agent receives as inputs the current map $m_T$ and the visual observation $o_T$ to predict a distant goal $g_T$ that will guide the robot to approach the viewpoint from which $I_g$ is taken.
\subsubsection{Learning Setup}
Before introducing our policy, we first describe the key ingredients of the learning setup, including datasets, inputs and output, and reward design.

\textbf{Datasets.} We conduct our learning on the Habitat simulator~\cite{savva2019habitat} with the visually realistic Gibson~\cite{xia2018gibson} dataset. We split the set of $86$ scenes from~\cite{xia2018gibson} into sets of $72/14$ scenes for training/testing, respectively. Each environment corresponds to a different apartment or house, generally including a kitchen, a bedroom, a living room, and a bathroom in the layout.
In the experiments, we further transfer the learned policies from Habitat to some real-world crowded scenes based on a mobile robotic platform (e.g., TurtleBot2). These scenes have never been encountered before.

\textbf{Inputs and output.} The long-term goal prediction policy takes a goal image, a current map, and a visual observation as inputs. At each time scale $T$, the goal image $I_g$ and the visual observation $o_T$ are both panoramas. Each panorama consists of four images, which are collected from $0^\circ$, $90^\circ$, $180^\circ$ and $270^\circ$ orientations of the agent, respectively.
The image resolution is $128\ast128$. The current map $m_T$ is drawn from our mapping module, which  consists of eight channels containing the obstacle area, the explored area, the current agent location and the past agent locations. Our model learns to analyze the connection between the current observation and the goal image and to exploit structural regularities of the current map to produce the possible position (namely, the long-term goal $g_T$) over the map where the goal image $I_g$ could be.

\textbf{Reward design.}
The objective of a reinforcement learning policy is to collect as many rewards as possible, and an informative reward function becomes a critical foundation on which a successful RL policy relies.
The reward function evaluates the behaviors drawn from the RL policy and always needs to be provided beforehand. However, in practice, defining the reward function can be challenging, since an informative reward function may be very difficult to specify and exhaustive to tune for large and complex problems~\cite{devo2020towards}.

Our purpose during the long-term prediction policy learning is to let an agent explore the environment effectively while moving towards the goal image and to avoid collisions during navigation.
Therefore, when the predicted long-term goal is near $I_g$, high values of reward are published. Collisions with obstacles are assigned penalties. In addition, to realize efficient navigation in unseen scenes, we encourage the agent to explore the environment before reaching the goal and penalize new explorations once $I_g$ is in the explored area of the agent.
We use the explored area increments to design the reward function, which results in faster convergence.
Formally, the total reward collected by the agent at time scale $T$ can be given as:

\vspace{-10pt}
\begin{equation}\label{eq:init_loss}
\begin{aligned}
r_T&=r_g+r_{collide}+r_{explore}(T)
\end{aligned}
\end{equation}

$r_g$ denotes the reward when the distance between the predicted long-term goal and the final goal $I_g$ is less than some threshold (e.g., $d_g=1.0m$).
$r_{collide}$ represents the penalty when the predicted long-term goal is located in the unreachable area of the environment. We set $r_g=20$ and $r_{collide}=-5$ in our formulation.
$r_{explore}(T)$ evaluates the agent exploration, which is given as:

\vspace{-10pt}
\begin{small}
\begin{equation}
r_{explore}(T)=\left\{
\begin{aligned}
&0               &&\textup{if}~T=0,\\
&\text{Exp}(T)-\text{Exp}(T+k) &&\textup{elif}~I_g \in Area(\text{Exp}(T)),\\
&\text{Exp}(T+k)-\text{Exp}(T) &&\textup{otherwise.}
\end{aligned}
\right.
\end{equation}
\end{small}
where $\text{Exp}(T)$ denotes the size of the explored area at time scale $T$;
$\text{Exp}(T+k)$ denotes the size of the explored area at the next time scale (namely, $T+k$);
and $I_g \in Area(\text{Exp}(T))$ represents $I_g$ is located in the explored area at time scale $T$.

\subsubsection{Model}
We focus on learning the long-term goal prediction policy function $\pi$ via deep reinforcement learning, where the long-term goal $g_T$ at time scale $T$ can be drawn by: 
\begin{equation}
g_T\sim \pi(I_g, m_T, o_T)
\end{equation}
The main task of the RL-based agent is to maximize the expected sum of future discounted rewards:
\vspace{-8pt}
\begin{equation}
\pi^{\ast}=arg\max_{\pi}E[\sum_{T=0}^{\infty}\tau^T r_{T}]
\end{equation}
where $r_T$ is the reactive reward provided by the environment at each time scale and  $\tau\in(0,1]$ is a discount factor.
We extend the Proximal Policy Optimization (PPO)~\cite{schulman2017proximal} to our parallel training framework. The policy is trained with experiences collected by all threads in parallel.
The parallel training framework not only dramatically reduces the time cost of
the sample collection but also makes the algorithm suitable for training in various environments.
Training in multiple environments simultaneously enables robust performances when generalizing to unseen scenes.

\begin{figure}[thpb]
\begin{center}
\includegraphics[width=\linewidth]{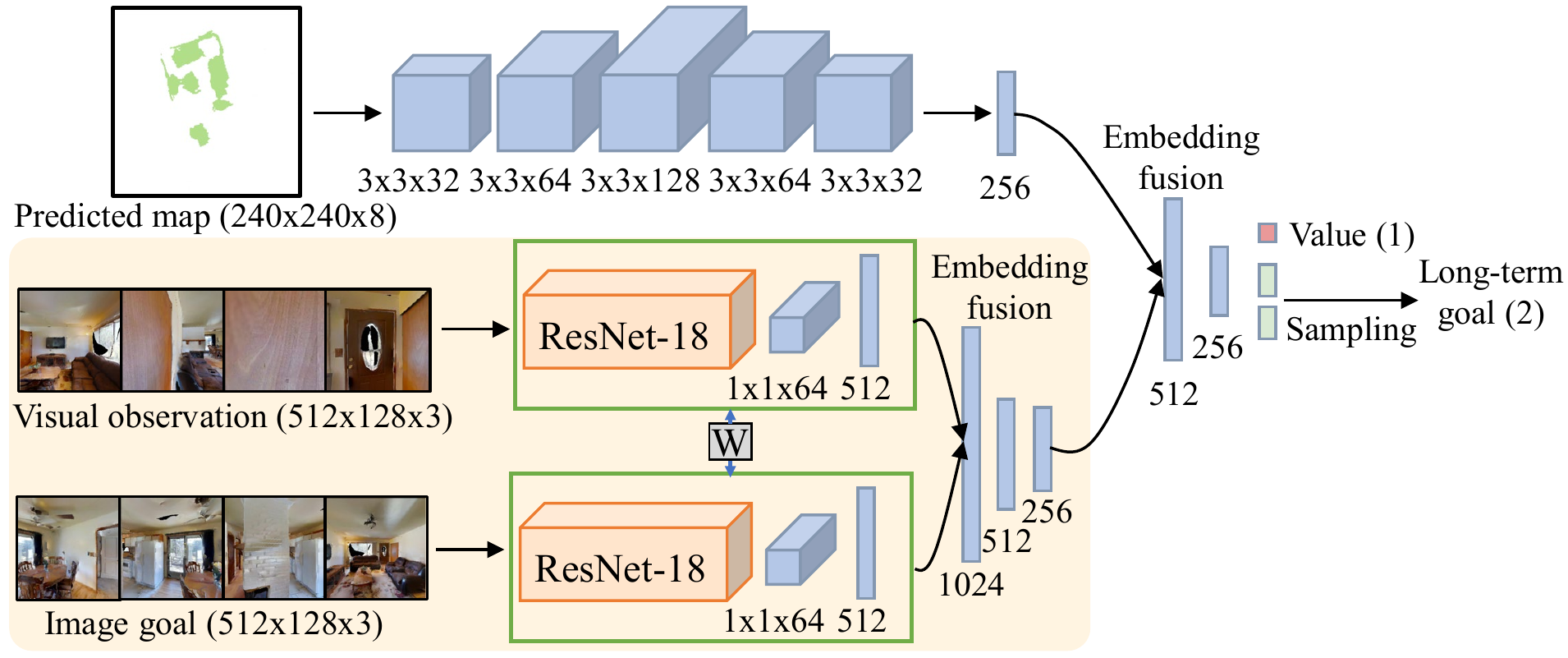}\vspace{-12pt}
\end{center}
\caption{The architecture of the long-term goal prediction module.}
\label{fig:long-term}\vspace{-10pt}
\end{figure}

\textbf{Network architecture.}
The long-term goal prediction policy demands an understanding of the relative spatial positions between the current observation and the goal image, as well as a holistic sense of the scene layout.
To reason about the spatial arrangement between the current observation $o_T$ and the goal image $I_g$, we use a two-stream deep siamese network for discriminative embedding learning. The backbone of the siamese layer is the ImageNet-pretrained ResNet-$18$~\cite{he2016deep} (truncated last layer) that produces a $512$-d feature on a $512\times128\times3$ RGB image.
The fusion layer takes a $1024$-d concatenated embedding of the state and the target, finally generating a $256$-d joint representation.
The current map $m_t$ is fed into five convolutional layers with $(32,64,128,64,32)$ filters with $3\times3$ kernel and stride $1$, all of them with rectified linear units (ReLU) as activation; it finally produces a $256$-d representation.
The two representations are concatenated together and processed by two fully connected layers,
and the second fully connected layer connects to three heads to output the value, the mean, and the variance of a Gaussian distribution, from which our $2$-d long-term goal $g_T$ is sampled.
The architecture is shown in Figure~\ref{fig:long-term}.

\textbf{Training details.} We train this network with an Adam optimizer of learning rate $lr=2.5e^{-6}$, exponential decay rates $betas=(0.9,0.999)$, and stability parameter $eps=(1e^{-6})$.
The training time is about $24$ hours on a high-performance computing cluster with two Intel Xeon Cascade Lake Gold $6248$ CPUs, $2.5$Hz, $20$ cores, and eight NVIDIA Tesla PCIE V$100$ GPUs.
During training, we estimate the discounted accumulative rewards and back-propagate through time for every $10$ unrolled time scales with $72$ navigation episodes executed at each time.
In addition, the number of navigation steps $k$ between two adjacent time scales is $10$.
Each episode terminates when the RL-based agent navigates $500$ time steps (namely, $50$ time scales) in an environment.


\subsection{The Reactive Motion Planning Module}
\label{sec:motion-planning}
The motion planning module takes as inputs the current sensor observation and the long-term goal $g_T$ at each navigation time step $(T+t),s.t.~t\leq k$ and outputs a navigational action.
This module is also decoupled from the other modules and any sensor-to-controls obstacle-avoidance agent could be used to take collision-free steering commands to reach the long-term goal $g_T$.
We explore this with two different motion planning technologies: the Fast Marching Method~\cite{sethian1996fast} and CrowdMove~\cite{fan2018crowdmove}. In Habitat, due to the lack of a laser sensor, we use the deterministic local motion planning algorithm, Fast Marching Method, to facilitate the learning of our long-term goal prediction module.
 An agent simply takes deterministic actions $\{a_{T+t}\}_{t=1}^{t=k}$ along the path to reach the long-term goal $g_T$, where $a_{T+t}\in \mathcal{A}$. $\mathcal{A}$ is a discrete set defined as: $\mathcal{A}=\{\emph{Move forward}; \emph{Turn right}; \emph{Turn left}\}$.
In real crowded scenarios,
we use CrowdMove to take collision-free steering commands to reach a long-term goal.
In our experiments, CrowdMove uses continuous actions for image-goal navigation, taking both robot dynamics and path feasibility into account and leading to better approximate optimal paths than the Fast Marching Method.
Note that although the goal prediction policy acts at a coarse time scale (e.g., $k=10$), the motion planning module acts at a fine time scale. At each time step $(T+t)$, we replan the motion $a_{T+t}$ to the long-term goal $g_T$ over the updated map $m_{T+t}$.

\subsection{The Robust Navigation Ending Prediction Module}
In this section, we train the navigation ending prediction module to issue a stop action at a correct location. With the module, an agent can figure out if a goal image is reached during the navigation in real unseen scenes.
The module takes as inputs the navigation goal image $I_g$ and the current visual observation $o_t$,
and outputs an ending indicator $e_t$ to determine whether the navigation is over.

This is essentially a binary classification problem and we design a siamese neural network like the one used in the long-term goal prediction module, including a pretrained ResNet-$18$ backbone and then a fusion layer (with the yellow background in Figure~\ref{fig:long-term}). The $256$-d feature from the fusion layer is then fed into a fully connected layer to produce the $2$-d ending indicator.
The navigation ending prediction module is jointly trained with our long-term goal prediction module in Habitat, although no parameters are shared between the two modules.
When the agent navigates in Habitat, we collect the navigation trajectories including the observation images and corresponding spatial positions. We construct the dataset for navigation ending prediction learning by selecting observation pairs randomly on these trajectories.
The spatial distance $d$ between two observations in a pair is used as a surrogate similarity measure. We define a label $e_{ij}^{gt}$ for each observation pair $(o_i,o_j)$ and the label $e_{ij}^{gt}$ is equal to $1$ if $d_{ij}\leq 1.0m$ and $0$ otherwise.
We train the navigation ending prediction module (NEPM) to predict the ending indicator $e_{ij}$ from the input pair $(o_i,o_j)$ with a binary cross-entropy loss as:

\begin{equation}\label{eq:target}
\begin{aligned}
\begin{split}
&l=E_{e_{ij}^{gt}\thicksim p(e|o_i,o_j)}[-\log{\textup{NEPM}(e_{ij}|o_i,o_j)}]\\
&s.t.~p(e|o_i,o_j)=\left\{\begin{matrix}
1 & e=e_{ij}^{gt} \\
0 & otherwise. \\
\end{matrix}\right.
\end{split}
\end{aligned}
\end{equation}

\emph{Training details.}
We update the network every two navigation time scales.
The batch size is set to $128$ for each back-propagation, and each batch has an equal number of positives and negatives for training data balance.
We also use Adam optimizer of learning rate $lr=2.5e^{-5}$, exponential decay rates $betas=(0.9,0.999)$, and stability parameter $eps=(1e^{-5})$.

%

%
%

%
%

%

\section{Experiments and Results}
\label{sec:Experiments_and_Results}
In this section, we evaluate our image-goal navigation performance on both simulated and real-world navigation tasks.

\subsection{Ablations and Baselines}
We first conduct experiments to analyze the effectiveness of different choices in our design.
As developed in Section~\ref{sec:Long-term}, our method includes a long-term goal prediction module, which allows an agent to explore the environment efficiently and finally determine the locations of image goals.
We first use a random policy (RP) to replace the module, denoted as \textbf{Ours-RP}.
The random policy randomly predicts a location on an online map at each time scale.
In addition, we ablate the effect of our proposed continuous rewards in Section~\ref{sec:Long-term} by using the sparse reward (SR) design in~\cite{zhu2017}, denoted as \textbf{Ours-SR}.
The motion planning module and the navigation ending prediction module are both crucial for our navigation method.
We exhibit the importance of the design over~\cite{chaplot2020neural}, which uses FMM~\cite{sethian1996fast} to convert a long-term goal to a short-term goal and trains an end-to-end local policy (LP) for both motion planning and navigation ending prediction. The local policy takes as inputs the current visual observation and the short-term goal and outputs a discrete action from $\mathcal{A}^{\ast}$. We use this design for an ablation and denote the variant as \textbf{Ours-LP}.

In addition, we compare our method with the following baselines:
(1)~\textbf{Random Agent (RA)} randomly picks an action from a discrete set of actions $\mathcal{A}^{\ast}$ at each time step, where $\mathcal{A}^{\ast}=\{\emph{Move forward}; \emph{Turn right}; \emph{Turn left};\emph{Stop}\}$.
(2)~\textbf{TD-VNM} represents the target-driven visual navigation model from~\cite{zhu2017} and is trained using standard reinforcement learning. We evaluate the navigation generalization to unknown scenes and thus we do not use the scene-specific design during the training.
(3)~\textbf{G-LSTM-RL-BC} is adapted from~\cite{wu2018building}, which incorporates a Gated-LSTM structure with an RL-based framework (e.g., A3C) and is trained using behavioral cloning (BC) to improve the navigation performance.
(4)\textbf{TD-RL-ITR} is an end-to-end image-goal visual navigation model~\cite{wu2021reinforcement} that designs an information-theoretic regularization to facilitate the RL policy leaning.
(5)~\textbf{NTS} is the abbreviation of `Neural Topological SLAM'~\cite{chaplot2020neural}, which studies the image-goal navigation in a hierarchy of three modules.

All the baselines are trained on Habitat. Please refer to Section~\ref{sec:Long-term} for the training data description. For evaluation, we sample navigation tasks in the testing scenes to create three levels of difficulty based on the distances between the image goal locations and the starting locations, as in~\cite{chaplot2020neural}: Easy $(1.5\sim3m)$, Medium $(3\sim5m)$, and Hard $(5\sim10m)$.
An agent succeeds in a navigation task if it predicts the stop action before the time limit (e.g., $500$ time steps) and the distance between the agent's current location and the goal image location is within a threshold (e.g., $1.0m$).
The agent fails if it takes stop action anywhere else or runs up to $500$ steps.
To measure the navigation performance, we use two metrics, success rate (SR) and success weighted by (normalized inverse) path length (SPL), as defined in~\cite{anderson2018evaluation}.
The higher the SR value, the better generalization, on average, the agent performs.
The higher the SPL value, the less search time, on average, the agent uses when approaching the target.

\subsection{Navigation in the Habitat Simulator}
In the Habitat simulator, our method uses pre-trained Neural-SLAM~\cite{chaplot2020learning} to maintain an online map at each time step during navigation, and then produces a long-term goal on the map at each time scale with the proposed goal prediction module. Next, our method uses the Fast Marching Method~\cite{sethian1996fast}, to plan a collision-free path to the long-term goal and uses the navigation ending prediction module to stop the navigation. We denote the whole architecture as \textbf{Ours}$_1$.
The architecture is similar to the variant of Active Neural SLAM model from~\cite{chaplot2020neural}, and the differences lie in the design of the long-term goal prediction module and the navigation ending prediction module. In addition, to avoid errors in pose prediction, we use the poses from the sensor of a navigation agent directly, leading to relatively noiseless maps. This is reasonable since our work is designed for image-goal navigation in real-world scenarios, and we always use Gmapping~\cite{grisetti2007improved} to provide the noiseless map of the real world. The mapping module is important, but not our main innovation. Our goal in this subsection is to evaluate the effectiveness of the proposed long-term goal prediction module during navigation. We analyze the navigation performances of the proposed method and all the baselines in the Habitat simulator.
The evaluation involves three difficulty levels, each containing $1000$ different navigation tasks randomly sampled from unknown scenes, as in~\cite{chaplot2020neural}.

\renewcommand{\multirowsetup}{\centering}
\begin{table}[h]
\centering
\caption{Average navigation performance (SR and SPL) comparisons on unseen scenes from the Habitat simulator.
\label{tab:table1}}\vspace{-10pt}
\scalebox{0.9}{\begin{tabular}{c|c|c|c|c}
\cline{1-5}
\hline
\multirow{2}{1.0cm}{Models}&Easy&Medium&Hard&Overall\\\cline{2-5}
    &SR / SPL &SR / SPL &SR / SPL &SR / SPL\\
\hline
Ours-RP&  0.65 / 0.32 & 0.51 /  0.28 &  0.39 / 0.21 & 0.51 / 0.27  	\\\cline{1-5}
Ours-SR&  0.37 / 0.21 & 0.23 /  0.10 &  0.09 / 0.01 & 0.23 /  0.11  \\\cline{1-5}
Ours-LP&  0.61 / 0.37 & 0.49 /  0.26 &  0.31 / 0.19 & 0.47 /  0.27  \\\cline{1-5}
Ours$_1$&  0.73 / 0.49 & \textbf{0.56} /  \textbf{0.33} &  \textbf{0.43} / \textbf{0.24} & \textbf{0.57} /  0.35 \\\cline{1-5}
\hline
Random Agent& 0.36 / 0.15 & 0.31 /  0.09 &  0.11 / 0.03& 0.26 /  0.09  	\\\cline{1-5}
TD-VNM~\cite{zhu2017}&  0.56 / 0.22& 0.17 /  0.06 &  0.06 / 0.02& 0.26 /  0.10  \\\cline{1-5}
G-LSTM-RL-BC~\cite{wu2018building}&  0.53 / 0.31 & 0.19 /  0.07 &  0.04 / 0.02 & 0.25 /  0.13  \\\cline{1-5}
TD-RL-ITR~\cite{wu2021reinforcement}&  0.68 / 0.36 & 0.45 /  0.23 &  0.19 / 0.09 & 0.44 / 0.22 \\\cline{1-5}
NTS~\cite{chaplot2020neural}&  \textbf{0.80} / \textbf{0.60}& 0.47 /  0.31 &  0.37 / 0.22 & 0.55 / \textbf{0.38}   \\\cline{1-5}
\hline
\end{tabular}}\vspace{-10pt}
\end{table}

\begin{figure}[thpb]
\begin{center}
\includegraphics[width=1.0\linewidth]{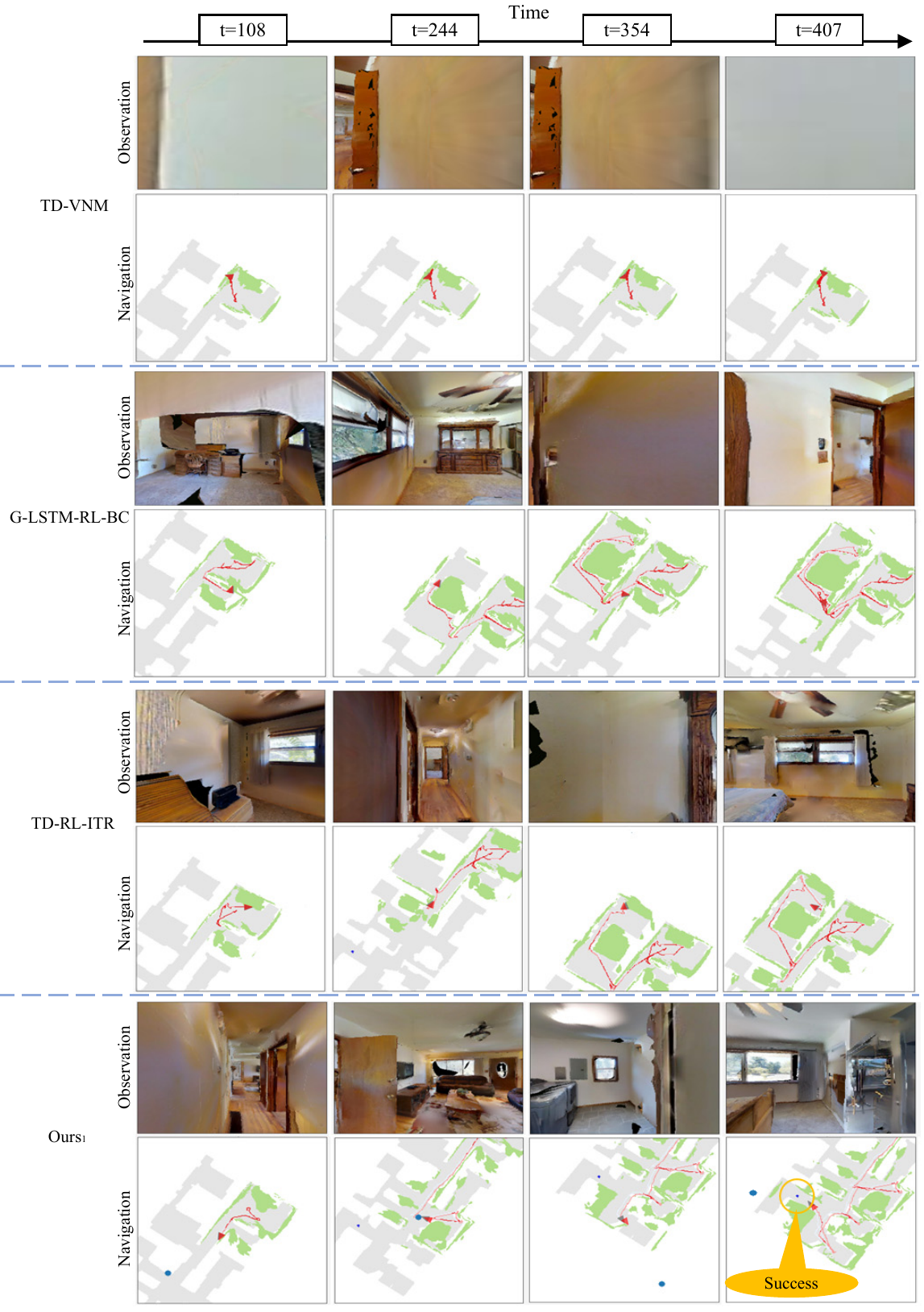}\vspace{-12pt}
\end{center}
\caption{Navigation visualization. We show trajectories of the proposed method and some baselines in a navigation task. Top: Front-view observations seen by the agent. Bottom: Local maps and trajectories. The ground-truth maps and poses are shown in grey. Trajectories generated by these navigation methods are shown in red. Map predictions from the Neural SLAM module are overlaid in green. Long-term goals selected by our goal prediction module are shown with large blue circles. The image goal positions are shown with small blue spots. Our agent can successfully stop in the yellow circle, which means the distance between the agent and the goal image is within $1.0m$.
}
\label{fig:tra}\vspace{-10pt}
\end{figure}

\subsubsection{Ablation}
We provide ablation results to gain insight into how navigation performances are affected by changing the structure. As shown in Table~\ref{tab:table1}, our navigation pipeline shows $6\%$ improvement in average SR, and $8\%$ improvement in average SPL over Ours-RP, which indicates the proposed long-term goal prediction module has a significant impact on improving the navigation capability of an agent.
Dealing with sparse rewards is especially challenging during navigation learning.
Ours-SR exhibits no learning even after millions of training frames, and it does not seem to follow any specific strategy during the evaluation.
The comparison results (Ours$_1$ vs. Ours-SR) suggest that our continuous reward design manages to explore the scene properly and guide the navigation efficiently.
Ours-LP considers motion planning and navigation ending prediction simultaneously,
and the performances of this ablation deteriorate as the difficulty of the navigation tasks increases.
Our model decouples the two processes and can tackle the hard level relatively well.


\subsubsection{Comparison}
Table~\ref{tab:table1} also summarizes the comparison results with some navigation baselines in the RGB setting. We compare with the state-of-the-art methods and show the generalization abilities of these navigation agents to transfer the learned navigation skills to previously unseen scenes.
Image-goal navigation is practically difficult and the general performances of these navigation agents are not very good, since the goal images can not directly provide the direction to explore compared to the point-goal navigation providing updated directions to the goals at each time step.
In addition, ending a navigation task in a right place is also challenging for these navigation agents, which requires analyzing the differences between the agent observations and the goal images.

In detail, the performances are intensely unfavorable in both the SR and the SPL metrics, when an agent applies random walk or is trained only by reinforcement learning as TD-VNM.
TD-VNM~\cite{zhu2017} originally designs different scene-specific layers for different scenes and thus the model lacks the generalization ability to unseen scenes.
We consider it is difficult to generate sensible navigation results for the pure RL-based agent due to the huge searching space.
G-LSTM-RL-BC~\cite{wu2018building} and TD-RL-ITR~\cite{wu2021reinforcement} are both end-to-end RL-based baselines and are trained with expert data in different ways.
As shown in Table \ref{tab:table1}, these two baselines cost much more searching time and have worse navigation performances (SR/SPL) than the hierarchical navigation agents (e.g., NTS~\cite{chaplot2020neural} and Ours$_1$).
Hence, it is more challenging for end-to-end RL networks to tackle the higher-order complicated control tasks.
On the other hand, since the proposed long-term goal prediction module can explore a new environment effectively, comparing with Neural Topological SLAM (NTS)~\cite{chaplot2020neural}, our method achieves better performances in both the Medium and Hard level navigation scenarios in the Habitat simulator as Table \ref{tab:table1}. However, the powerful exploration can be troublesome when the navigation ending prediction module loses confidence, which leads to a little worse performances in the Easy scenarios.


Figure~\ref{fig:tra} visualizes the trajectories of the proposed method and some baselines for a navigation task. Note that we also present the automatic mapping results for these compared baselines.
As can be seen, the agent based on TD-VNM~\cite{zhu2017} gets stuck in the corner and thus the mapping area is small. The agents based on G-LSTM-RL-BC~\cite{wu2018building} and TD-RL-ITR~\cite{wu2021reinforcement} both present limited exploration capabilities and waste much time in exploring similar areas, and finally fail to approach the goal image. Our method explores the scene effectively and successfully navigates the agent to the goal.

\vspace{-8pt}
\renewcommand{\multirowsetup}{\centering}
\begin{table}[h]
\centering
\caption{Average navigation performance (SR and SPL) comparisons on unseen scenes from the real world.
\label{tab:table3}}\vspace{-10pt}
\scalebox{0.85}{\begin{tabular}{c|c|c|c|c|c}
\cline{1-6}
\hline
\multirow{2}{1.0cm}{Models}&(b)&(c)&(d)&(e) &(f)\\\cline{2-6}
 &SR/CR &SR/CR &SR/CR &SR/CR &SR/CR \\
\hline
TD-VNM~\cite{zhu2017}&  0.14/0.46 & 0.06/0.60 &  0.24/0.36 & 0.10/0.50  & 0.12/0.48	\\\cline{1-6}
G-LSTM-RL-BC~\cite{wu2018building}&  0.18/0.42& 0.08/0.52 &  0.20/0.34& 0.06/0.46& 0.08/0.42  \\\cline{1-6}
TD-RL-ITR~\cite{wu2021reinforcement}&  0.28/0.38 & 0.22/0.54 &  0.32/0.30& 0.14/0.38& 0.18/0.34\\\cline{1-6}
Ours$_2$&  \textbf{0.50}/\textbf{0.14}& \textbf{0.32}/\textbf{0.20} &  \textbf{0.58}/\textbf{0.10}& \textbf{0.26}/\textbf{0.20} & \textbf{0.34}/\textbf{0.16}\\\cline{1-6}
\hline
\end{tabular}}\vspace{-10pt}
\end{table}

\subsection{Navigation in the Real World}
\begin{figure}[thpb]
\begin{center}
\includegraphics[width=\linewidth]{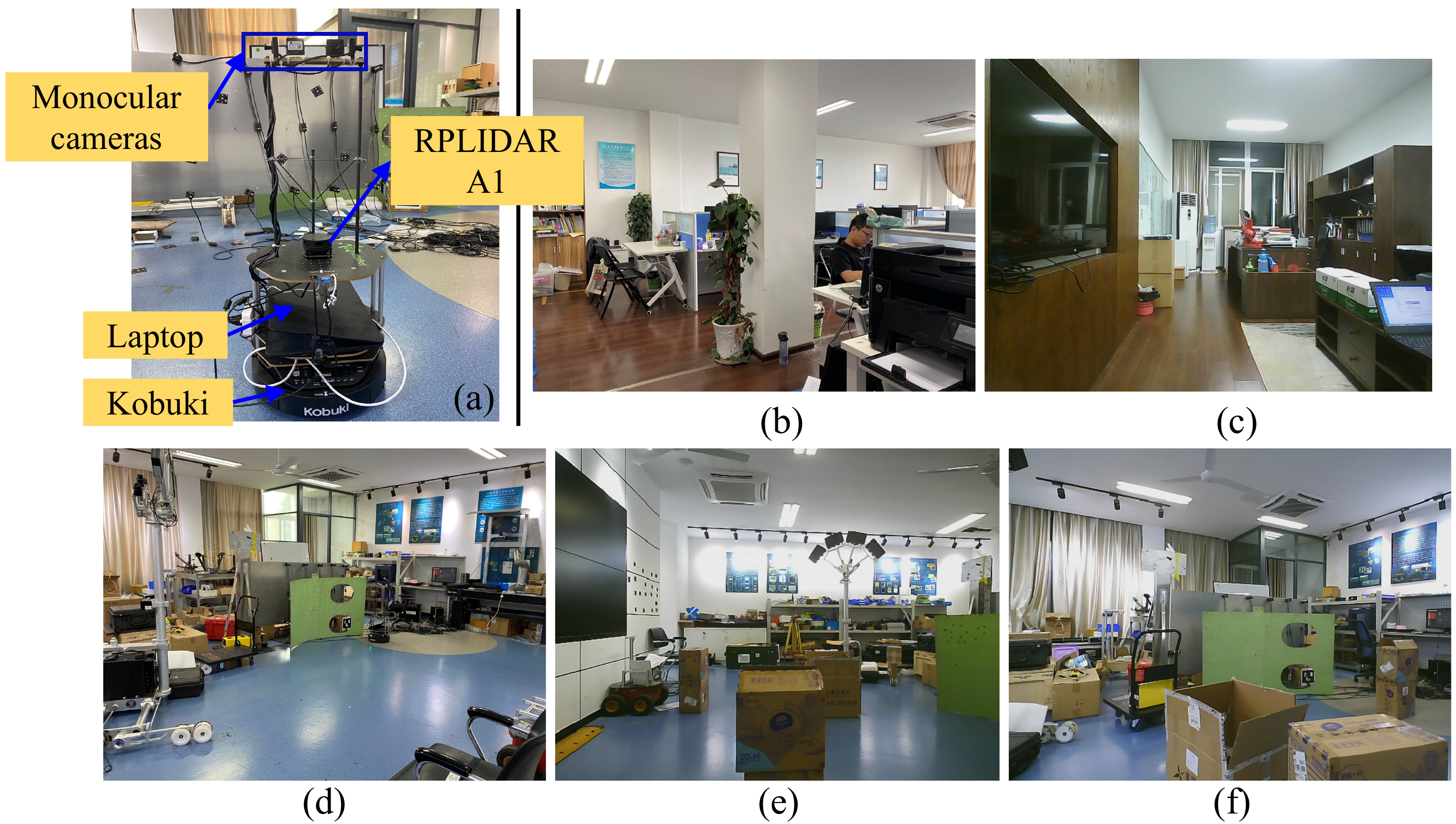}\vspace{-12pt}
\end{center}
\caption{(a) The robot system setup. (b-f) The physical indoor environments with different configurations.}
\label{fig:robo}\vspace{-10pt}
\end{figure}

We demonstrate the proposed navigation system using the TurtleBot2 robot. The hardware configuration of TurtleBot2 is shown in Figure~\ref{fig:robo}(a), which consists of a Kobuki base, an onboard laptop, an RPLIDAR A1 laser, and four monocular cameras.  The moving base executes the steering command output from the navigation system, which is deployed on the laptop.
The laser sensor and the camera sensors provide the inputs to the proposed system.
The output space of the system is a set of permissible velocities in continuous space, as described in Section III-C.
The navigation experiments are conducted on three physical indoor environments with five different configurations, as shown in Figure~\ref{fig:robo}(b-f).

We first deploy \textbf{Ours}$_1$ on the laptop and visualize the mapping results, which are inferior and lead to poor navigation performances. We consider that Neural-SLAM~\cite{chaplot2020learning} is sensitive to the robot setting and can not bridge the simulation to reality gap, especially when the real robot setting is as simple as ours.
To tackle the image-goal navigation in real settings, as mentioned in Section~\ref{sec:Hierarchical}, our method uses Gmapping~\cite{grisetti2007improved} to update a navigation map in real time and uses the trained long-term goal prediction module to produce a long-term goal on the map at each time scale.
In addition, CrowdMove~\cite{fan2018crowdmove} is exploited for the motion planning to the long-term goal and uses continuous actions to provide better approximate optimal paths than the Fast Marching Method~\cite{sethian1996fast}.
Finally, the navigation ending prediction module stops the navigation agent at a proper location.
We denote the whole architecture as \textbf{Ours}$_2$.
In addition, all the tested navigation models are trained purely in simulation, and the real-world environments are unknown to these navigation agents. We test the image-goal navigation capabilities, including the average success rate and the collision rate (CR, the rate of collision cases to all navigation cases).
The collision rate is measured to evaluate the robustness of navigation models to real sensor noises and thus is not provided during the simulation. The robot and the goal image are both placed randomly in these environments and the objective of the robot is to approach the goal image as fast as possible. We evaluate $50$ navigation tasks in each configuration.
The authors of~\cite{chaplot2020neural} do not provide the code and the navigation results from the real world, so we do not compare with this work here.
We compare our method with TD-VNM~\cite{zhu2017}, G-LSTM-RL-BC~\cite{wu2018building}, and TD-RL-ITR~\cite{wu2021reinforcement}.
The quantitative analysis of the navigation performances is provided in Table~\ref{tab:table3}.

\begin{figure}[thpb]
\begin{center}
\includegraphics[width=\linewidth]{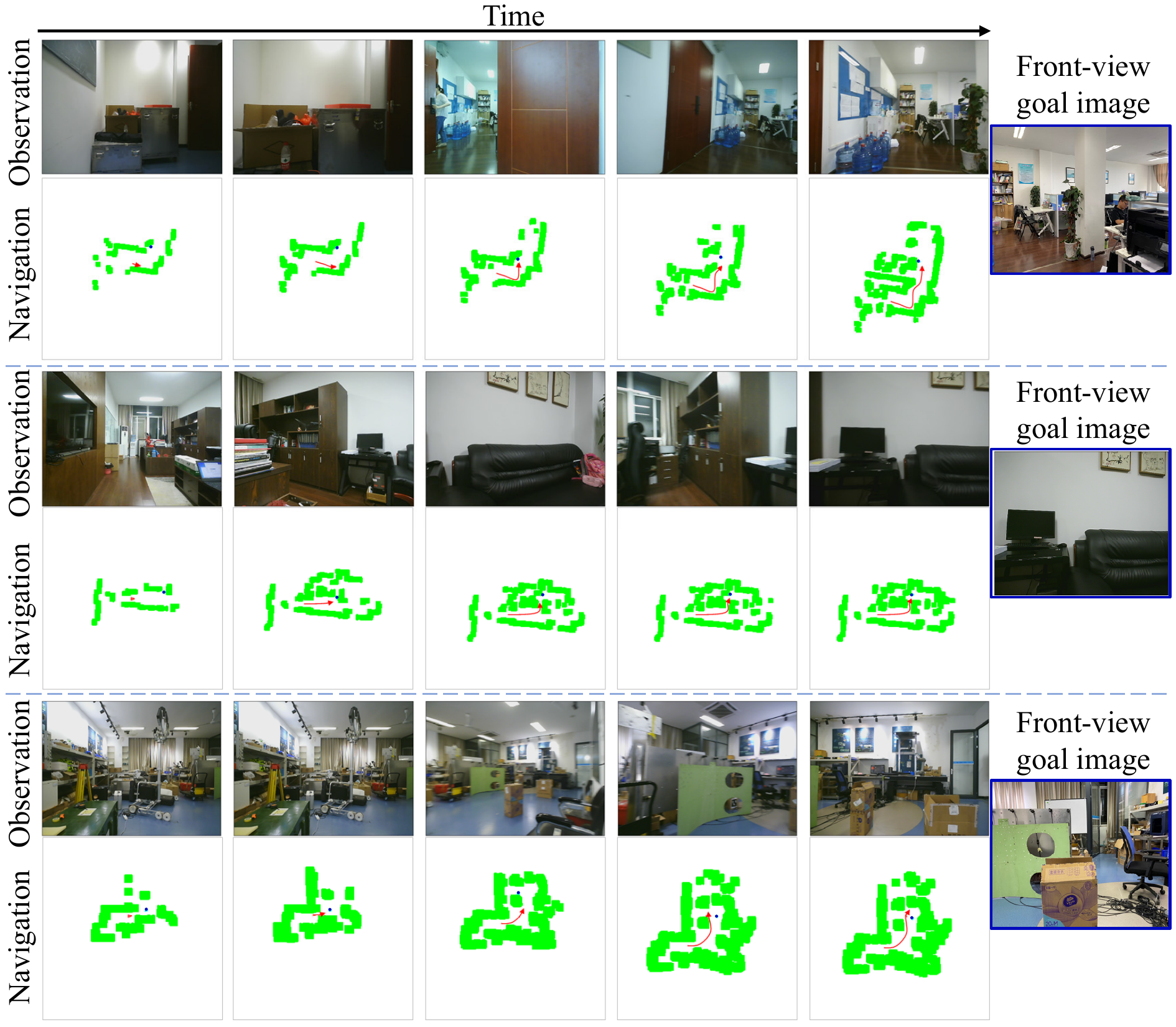}\vspace{-12pt}
\end{center}
\caption{Real-world transfer. We show three successful trajectories of the proposed method transferred to the real world. Sample images seen by the robot are shown on the top and the local maps in green and the trajectories in red are shown below. The long-term goals selected by the goal prediction module are shown in blue.}
\label{fig:real_tra}\vspace{-10pt}
\end{figure}

In the experiments, the pure RL-based model (e.g., TD-VNM~\cite{zhu2017}) tends to be more cautious, and it thrashes around in space without making progress, resulting in a lower success rate.
G-LSTM-RL-BC~\cite{wu2018building} and TD-RL-ITR~\cite{wu2021reinforcement} are both end-to-end navigation models, which are both trained with supervision from the shortest paths of navigation tasks.
The two models collide more often because learning about navigation and collision avoidance simultaneously is especially difficult for the end-to-end training strategy.
As expected, Ours$_2$ generalizes well to the unseen real-world environments and shows the best navigation performances in both success rate and collision rate.
This can be explained by the fact that we decouple the learning of image-goal navigation and collision avoidance by designing two mutually independent modules; thus, each module can demonstrate more concentrated learning capabilities.
In addition, the three navigation baselines all predict discrete action commands, leading to significantly oscillatory and jerky motions during robot navigation.
Using our proposed navigation system, the mobile robot can move continuously in most cases. This is an important navigation property in a physical environment.

In practice, we also found that the navigation performances are highly affected by the complexity of navigation tasks (e.g., Table~\ref{tab:table3} (3) vs. Table~\ref{tab:table3} (4)).
Complex navigation tasks generally lead to poor navigation performances.
In our case, the long-term goal prediction module and the navigation ending prediction module are the main problems and they are trained in the Habitat simulator and directly used in the real scenarios.
The large differences between the reality and the simulation seriously affect the two modules.
In addition, the reactive motion planning module is partly responsible, which leads to some occurrences of the robot freezing and collisions during navigation due to the sensor noises and the actuation delays of our robot.
Hence, exploring more effective motion planning methods will be helpful.
Figure~\ref{fig:real_tra} visualizes three trajectories of TurtleBot generated by our system. Please see the video in the supplementary material for additional results obtained from the experiment in the real world.

\section{Conclusion}
\label{sec:conclu}
In this work, we proposed a novel navigation method to tackle the image-goal navigation in real complex environments.
The core of the method is hierarchical decomposition and modular learning, which decouples the learning of navigation planning, collision avoidance, and navigation ending prediction.
This enables more concentrated learning and helps achieve state-of-the-art performances on the image-goal navigation.
We show that a direct sim-to-real transfer is possible. The proposed hierarchical framework generalizes well in real crowded scenarios.
However, although the results are promising, there are still some open problems.
Both the reliability and the robustness of the proposed modules including the navigation goal planning, the motion planning, and the ending prediction, should be improved.
In future work, we will explore effective techniques to improve overall performances. For example, semantic properties of environmental objects can be used during the navigation goal planning, as in~\cite{chaplot2020object}.
Incorporating $3$D environmental perception~\cite{zhu2018scores,qi2019deep} for better generalization is also a great topic for future work.

\bibliographystyle{IEEEtran}

\bibliography{bibtex}
\ifCLASSOPTIONcaptionsoff
  \newpage
\fi

\end{document}